\pdfoutput=1

\documentclass[11pt]{article}

\usepackage[final]{acl}

\usepackage{times}
\usepackage{latexsym}
\usepackage{amsmath}
\usepackage{booktabs}
\usepackage{float}
\usepackage{xcolor}
\usepackage{tcolorbox}
\usepackage{paralist}

\usepackage[T1]{fontenc}

\usepackage[utf8]{inputenc}

\usepackage{microtype}

\usepackage{inconsolata}

\usepackage{graphicx}
\title{Rationalizing Transformer Predictions via End-To-End Differentiable Self-Training}

\author{Marc Brinner \and Sina Zarrieß \\
  Computational Linguistics, Department of Linguistics\\
  Bielefeld University, Germany\\
  \texttt{\{marc.brinner,sina.zarriess\}@uni-bielefeld.de}}

\begin{document}
\makeatletter
\def\ps@firstpage{
  \def\@oddfoot{
    \hfil\parbox{\textwidth}{
      \centering
      \thepage\\[4pt] 
      \small{\textit{Proceedings of the 2024 Conference on Empirical Methods in Natural Language Processing}, pages 11894--11907\\[-0pt] 
      November 12-16, 2024 \copyright2024 Association for Computational Linguistics}
    }
    \hfil
  }
}

\def\ps@plain{
  \def\@oddfoot{
    \hfil\parbox{\textwidth}{
      \centering
      \thepage\\[4pt]
      \vspace{20px}
    }
    \hfil
  }
}
\makeatother

\pagestyle{plain}
\setcounter{page}{11894}
\maketitle
\thispagestyle{firstpage}
\begin{abstract}

We propose an end-to-end differentiable training paradigm for stable training of a rationalized transformer classifier. Our approach results in a single model that simultaneously classifies a sample and scores input tokens based on their relevance to the classification. To this end, we build on the widely-used three-player-game for training rationalized models, which typically relies on training a rationale selector, a classifier and a complement classifier. We simplify this approach by making a single model fulfill all three roles, leading to a more efficient training paradigm that is not susceptible to the common training instabilities that plague existing approaches. Further, we extend this paradigm to produce class-wise rationales while incorporating recent advances in parameterizing and regularizing the resulting rationales, thus leading to substantially improved and state-of-the-art alignment with human annotations without any explicit supervision.
\end{abstract}
\section{Introduction}

Neural networks are increasingly prevalent across a wide range of applications, driving significant advancements in fields such as natural language processing, computer vision, and beyond. Due to the black-box nature of these networks, this widespread use comes with an increased demand for interpretability \cite{lyu2024towards}, as understanding the basis for the decisions made by these models is crucial for their reliable and ethical deployment. This need has become especially clear with the increasing use of notoriously uninterpretable large language models, which have the potential to quickly lose a user's trust after only few confidently incorrect predictions \cite{dhuliawala-etal-2023-diachronic}.

One possible mitigation is the use of encoder-only models, which lend themselves more readily to classical interpretability approaches designed for general neural network classifiers while still providing state-of-the-art performance due to continuous improvements in model structure \cite{he2021deberta, he2023debertav3} and training paradigms \cite{zhang2023veco}.

While a variety of explainability methods exist, that usually assign scores to input tokens indicating their importance for a classification \cite{sun2021interpreting}, these methods often suffer from several drawbacks, including high computational cost, difficult-to-interpret explanations, and potentially even unfaithful representations of the model's decision-making process. In this study, we close this gap by developing a rationalized transformer predictor that generates faithful and interpretable explanations in addition to its decisions within the same forward pass.

As a foundation for our approach, we build upon the existing and commonly used three-player game proposed by \citet{yu2019rethink}. In this framework, a selector model chooses a subset of the input as rationale, while a predictor and a complement predictor model are trained to infer the correct label from either the tokens included in the rationale or the tokens not included in the rationale, respectively. The selector model is then trained to maximally aid the predictor in predicting the correct label while preventing the complement predictor from doing the same, thus ensuring that all tokens indicative of the correct label are included in the rationale.

While the general three-player game is sensible, the actual realizations that are proposed often have several limitations, including being not end-to-end differentiable due to a stochastic sampling process in the forward pass, showing interlocking dynamics that might prevent convergence to a suitable solution, and having no guarantee of providing a rationale that actually explains the prediction (compare Section \ref{sec:issues} for a more detailed discussion).

For this reason, we propose a new take on this three-player game that is not susceptible to these drawbacks. We achieve this by making use of a single unified model that is trained as a standard classifier on the complete unaltered input, while simultaneously predicting class-wise importance scores for each input token in the same forward pass, which are then trained using self-training to mark spans that the model itself considers important for the specific class.

Our proposed rationalized transformer predictor (RTP) simplifies and enhances the common three-player structure in several ways, including 1) using only a single model to fulfill all three roles of the three-player game, thus enabling classification and rationale prediction in a single forward pass 2) training the rationales to explain the predictor, but avoiding training the predictor on the rationales, which ensures that the rationales faithfully explain the predictions 3) creating rationalized inputs in continuous fashion to enable fully differentiable training and avoid sampling 4) creating class-wise rationales and 5) using a parameterization that maximizes similarity with human rationale annotations.

We evaluate our method on two benchmarks for explainable AI and compare it with existing post-hoc explanation methods as well as methods leveraging standard multi-player procedures. We show that our method achieves state-of-the-art performance on both tasks, demonstrating previously unseen alignment with human rationales in combination with high rationale faithfulness.

\section{Background}

\subsection{Post-Hoc Rationalization}

Since neural networks are black-box models, the ever-increasing use of such models in research and industry has led to a strong demand for methods that reliably explain neural network classifications. To this end, a variety of approaches have been proposed, many of which are designed to create post-hoc explanations for an already trained classifier. These methods rely on a variety of mechanisms, including 1) making use of the models gradients at different inputs to obtain importance scores \cite{simonyan2013deep, sundararajan2017intgr} 2) quantifying the influence of individual input elements by observing the effect of input perturbations on the predicted outputs \cite{castro2009shapley,zeiler2014vis, zhou2014detect,vitali2018rise} 3) fitting interpretable models to neural-network outputs \cite{riberio2016lime} 4) developing backpropagation-like procedures to propagate importance information from the model output to the input features \cite{zeiler2014vis, spring2015simpl, bach2015relevance, shrikumar2017deeplift, Chefer2021generic, Chefer2021transformer} 5) performing input-optimization to create an altered input that only retains the information important for the classification \cite{brinner2023MaRC}.

\subsection{Rationalized Classification}

Due to the inherent difficulty of creating post-hoc explanations for classifiers that were never designed to be explainable, rationalized predictors have been proposed that are explicitly trained to perform the original task while simultaneously providing a rationale for the prediction in a single forward pass. \citet{lei2016rational} were the first to propose a two-player game for textual inputs, involving a rationale selector model and a classifier model. The rationale selector assigns a probability to each input word, indicating its likelihood of belonging to the rationale, so that a discrete rationale can be sampled from this distribution. The classifier then uses only the rationale to make its classification, thus ensuring that the selected words were responsible for the classification. During training, the classifier is trained as usual to predict the correct label from a sampled rationale, while the rationale selector is trained to produce rationales that aid the classifier in making the correct predictions, ensuring that words indicative of the correct class are selected.

\subsection{Common Issues of Rationalized Classifiers}
\label{sec:issues}
While the general training paradigm of the two-player game is sensible, several issues affect the training, performance and faithfulness of the rationales:
\begin{enumerate}
    \itemsep-3px
    \item \textbf{Stochastic Sampling}: Training requires stochastic sampling of rationales, meaning that gradients can only be estimated using methods like REINFORCE \cite{williams1992reinforce}, which are generally less stable and slow to convergence.
    \item \textbf{Class-Independent Rationales}: A single rationale is predicted regardless of the sample's class. In case a sample belongs to multiple classes, it is not possible to identify which part of the input is indicative of a specific class.
    \item \textbf{Interlocking Dynamics}: Interlocking dynamics might lead to degenerate solutions, for example, if the rationale predictor adapts too quickly to the noisy rationales that are produced by the randomly initialized rationale selector or vice versa \cite{yu2021interlock}.
    \item \textbf{Dominant Selector}: The training paradigm enforces rationales that persuade the classifier to predict a label that the predictor deemed correct, which does not necessarily correspond to faithful explanations of the actual reasoning process \cite{jacovi2021align}. In extreme cases, the rationale generator might simply encode the correct classification in the rationale (e.g., by selecting a specific kind of token), so that the classifier does not perform any significant reasoning itself.
    \item \textbf{Mismatch with Human Annotations}: Often, rationales are most useful if they resemble rationales provided by human annotators. Despite regularizers designed to enforce the selection of longer, consecutive spans of text, models often struggle to select spans that match human annotations, since overly strong regularization often overpowers the weak gradient signal created by REINFORCE, leading to degenerate solutions (e.g., selecting no tokens or all tokens).
    \item \textbf{Degraded Classification Performance}: The actual classification performance often degrades compared to standard classifiers \cite{jacovi2021align}.
\end{enumerate}

Several approaches have been proposed to modify or extend the two-player game to address these issues. \cite{liu2022fr} address the dominant selector issue by using a shared encoder for both the selector and the classifier, thus ensuring that both components focus on similar features instead of, in the case of a dominant selector, an encoded message. \citet{yu2019rethink} instead extended the two-player paradigm into a three-player game by introducing a complement predictor that is trained to predict the correct label from all words not included in the rationale. The rationale selector is then trained to prevent the complement predictor from identifying the correct class, thus ensuring that all words indicative of the correct class are selected as rationale, addressing the interlocking problem and (in part) the problem of having a dominant selector. \citet{chang2019classwise} propose the CAR framework that uses two encoders and one decoder per class to generate class-wise (and potentially counterfactual) rationales, solving issues 2, 3 and (in part) issue 4. They also use the straight-through gradient estimator \cite{Bengio2013EstimatingOP} instead of using REINFORCE, which addresses issue 1. \citet{liu2023multi} make use of multiple generators to mitigate issue 3, while the A2R method \cite{yu2021interlock} addresses the same issue by introducing a separate predictor that uses a soft selection of inputs instead of binary thresholding. To our knowledge, our proposed method for training a rationalized classifier is the only one to address all of the issues discussed above.

\section{Method}
\begin{figure*}
    \centering
     \includegraphics[width=\textwidth]{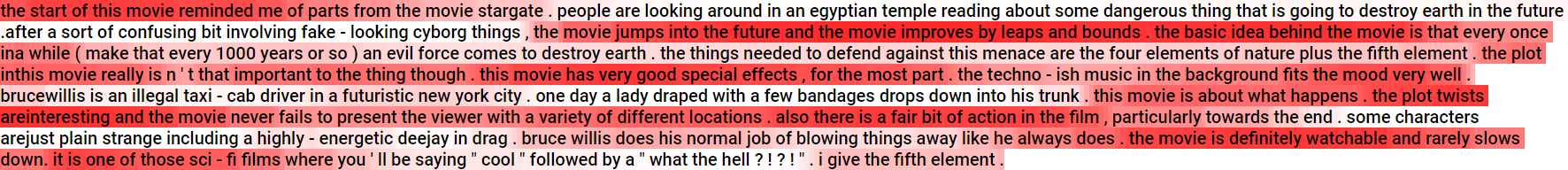}
    \caption{An exemplary output of the RTP for a positive review from the movie reviews dataset.}
    \label{fig:preiction}
    \vspace{-9px}
\end{figure*}
We propose a new method for end-to-end differentiable training of a rationalized transformer predictor (RTP). In the following, plain letters (e.g., $x$) denote scalars, while bold letters (e.g., $\textbf{x}$) denote vectors or tensors. We assume a text classification problem with label set $\mathcal{Y}$, and a training set consisting of texts $\textbf{x}_0,...,\textbf{x}_n$ with corresponding ground truth vectors $\textbf{y}_0,...,\textbf{y}_n$.

\subsection{Concept}
The RTP relies on a single model that, in one forward pass, produces both a classification output and class-wise importance scores for each token, denoting how indicative each token is of the respective class. The classification component is trained as a standard classifier, while the token-wise rationales are trained by creating altered inputs that only retain the important information for each individual class. The quality of these altered inputs (and therefore the quality of the rationales) is judged by the model itself by passing them through the model and observing its classification output. Through this end-to-end differentiable procedure, the rationales are optimized to faithfully explain the model predictions.

\subsection{Model Structure}
The basis of our method is a single model $M$, that, given an input text $\textbf{x}$, simultaneously predicts class probabilities $\tilde{\textbf{y}}$, as well as a mask tensor $\textbf{m}$:
\begin{align}
    \tilde{\textbf{y}}, \textbf{m} = M(\textbf{x})
\end{align}
with the mask $\textbf{m}$ being the rationale for the classification output $\tilde{\textbf{y}}$. Notably, $\textbf{m}$ consists of $|\mathcal{Y}|$ individual vectors $\textbf{m}^0,...,\textbf{m}^{|\mathcal{Y}|-1}$ that constitute individual rationales for each class $c\in\mathcal{Y}$, with each $\textbf{m}^c$ being a vector containing a mask value $m^c_i$ in the range $0$ to $1$ for each input token $x_i$, indicating its influence on the predicted likelihood of class $c$. In practice, the basis for classification output $\tilde{\textbf{y}}$ will be the \textit{CLS}-token embedding of the transformer classifier, while the mask values $\textbf{m}$ will be calculated from the predicted outputs for each token.

\subsection{Mask Parameterization}
In this section, we will discuss the parameterization that transforms token-wise neural network outputs into a smooth mask. The RTP outputs a mask for each individual class, but since mask calculations for individual classes are independent of each other, we will look at the mask $\textbf{m}^c$ for a single class $c$, which we denote as $\textbf{m}$ for simplicity.

A simple mask parameterization would predict a logit $l_i$ for each token $x_i$ and define $m_i=\sigma(l_i)$. Even with regularizers that enforce smooth mask selections, this approach often fails to select long spans of text as rationales, which would be desirable for matching human annotations. For this reason, we opted for the mask parameterization proposed by \citet{brinner2023MaRC} that explicitly enforces the prediction of longer spans of text as rationales by letting neighboring mask values influence each other.

In this parameterization, the model outputs two values $w_i$ and $\sigma_i$ for each word $x_i$. $w_i$ is mainly responsible for determining the mask value of word $x_i$, while $\sigma_i$ determines the influence of $w_i$ on the mask values of neighboring words. Introducing regularizers to enforce large values for $\sigma_i$ then leads to smooth masks. The mathematical formulation of the parameterization is as follows:
\begin{align}
    w_{i \rightarrow j} = w_i \cdot \exp\big(-\frac{d(i, j)^2}{\sigma_i}\big) \\
    m_j = \textrm{sigmoid}(\sum_i w_{i \rightarrow j})
\end{align}
Here, $d(i, j)$ denotes the distance between two words $x_i$ and $x_j$ and $w_{i \rightarrow j}$ is the influence of $w_i$ on the mask value of word $j$. $m_j$ is then calculated by applying the sigmoid to the sum of all influence values, resulting in the mask $\textbf{m}$ for the specific class at hand. Predicting masks $\textbf{m}^0,...,\textbf{m}^{|\mathcal{Y}|-1}$ for each class will simply be done by predicting individual outputs $\textbf{w}^c$ and $\boldsymbol{\sigma}^c$ for each class $c$ and performing the calculations independently.

\subsection{Model Training}

Given a sample $(\textbf{x}, \textbf{y})$, the classification capabilities of model $M$ are trained like a standard neural network classifier by performing a prediction and applying a loss function like cross-entropy loss to the predicted output. In contrast to other rationalized models, our training paradigm therefore trains the classifier on the unaltered input, not on a masked version that might remove crucial information.

To train the rationale predictions (i.e., the masks $\textbf{m}$), we use the current mask predictions to create two altered inputs $\textbf{x}^c$ and $\overline{\textbf{x}}^c$ for each ground-truth class $c$, with input $\textbf{x}^c$ retaining all information that is indicative of class $c$ according to mask $\textbf{m}^c$, while $\overline{\textbf{x}}^c$ is the complement input that removes all information specified by mask $\textbf{m}^c$:
\begin{align}
    \textbf{x}^c = \textbf{m}^c \cdot \textbf{x} + (1-\textbf{m}^c) \cdot b \\ \overline{\textbf{x}}^c = (1-\textbf{m}^c) \cdot \textbf{x} + \textbf{m}^c \cdot b
\end{align}
Here, $b$ denotes an uninformative background (e.g., \textit{PAD}-token embeddings). Notably, the mask $\textbf{m}$ is applied in continuous fashion to $\textbf{x}$ and $b$, meaning that embeddings for words are linearly blended towards uninformative embeddings according to $\textbf{m}$. In contrast to sampling of a discrete mask, this ensures full differentiability and was proven to have the desired effect of gradual removal of information by \citet{brinner2023MaRC}.

The rationalized inputs are then fed back into the same model $M$ and are scored by its classification component. We then make use of a loss function that rewards predicting the correct label from $\textbf{x}^c$, but not from $\overline{\textbf{x}}^c$, meaning that all information indicative of class $c$ is contained in the rationale (thus enforcing rationale comprehensiveness). The loss formulations used are the following:
\begin{align}
    \mathcal{L}^c = \textbf{CE}(M(\textbf{x}^c), \textbf{y})\\
    \mathcal{L}^{\overline{c}} = \textrm{relu}(M(\overline{\textbf{x}}^c)[c] - \alpha)
\end{align}
where \textbf{CE} denotes the cross-entropy loss, $M(\textbf{x})[c]$ denotes the predicted probability of class $c$ and $\alpha$ is a hyperparameter ensuring that the model is not required to drive the probability of class $c$ for input $\overline{\textbf{x}}^c$ to $0$, since driving it to a small value is sufficient. Importantly, the model $M$ is only trained with respect to the first forward pass that produces the rationales, and is therefore not updated to improve classification on the altered inputs $\textbf{x}^c$ and $\overline{\textbf{x}}^c$. This ensures that the classification performance in not influenced, and that the rationales actually explain the classification instead of dictating it. The final optimization problem looks as follows:
\begin{align}
\label{eq:objective}
\begin{split}
    \underset{M}{\textrm{arg\,min}}\:\:\: \textbf{CE}(M(\textbf{x}),& \textbf{y}) + \sum_{c\in\textbf{y}}\left[\mathcal{L}^c + \mathcal{L}^{\overline{c}}\right]\\&+  \Omega_{\lambda} + \Omega_\sigma
\end{split}
\end{align}
where $c\in\textbf{y}$ indicates summing over all ground-truth labels and $\Omega_{\lambda}$ and $\Omega_\sigma$ denote regularizers that enforce sparsity and smoothness of the rationales, respectively. Details on these regularizers and further details on the general objective are available in Appendix \ref{app:obj_details}.

\begin{table*}[t]
  \centering
  \fontsize{9.5pt}{10.8pt}\selectfont
  \begin{tabular}{lccccccccc}
    \toprule
    Method  & Clf-F1 & AUC-PR & Token-F1 & D-Token-F1 & IoU-F1 & D-IoU-F1& Suff. $\downarrow$ & Comp.$\uparrow$ & Perf. \\
    \midrule
    Random & - & 0.220 & 0.255 & 0.222 & 0.067 & 0.003 & 0.194 & 0.191 & 0.289 \\
    Supervised & 0.730 & 0.557 & 0.406 & 0.509 & 0.231 & 0.257 & 0.005 & 0.396 & 1.028 \\
    \midrule
    MaRC & \underline{0.776} & 0.366 & 0.336 & 0.351 & \underline{0.219} & \underline{0.178} & 0.040 & 0.459 & 0.974 \\
    Occlusion & \underline{0.776} & 0.307 & 0.277 & 0.294 & 0.145 & 0.071 & 0.078 & 0.352 & 0.696 \\
    Int. Grads & \underline{0.776} & 0.315 & 0.302 & 0.318 & 0.087 & 0.013 & 0.030 & 0.538 & 0.897 \\
    LIME & \underline{0.776} & 0.272 & 0.280 & 0.273 & 0.082 & 0.007 & 0.097 & 0.406 & 0.671 \\
    Shapley & \underline{0.776} & 0.309 & 0.301 & 0.320 & 0.084 & 0.009 & \underline{-0.012} & \underline{0.587} & 0.984 \\
    L2E-MaRC & \underline{0.776} & \underline{0.431} & \underline{0.359} & \underline{0.402} & 0.174 & 0.131 & 0.044 & 0.503 & \underline{0.992} \\
    \midrule
    2-Player & 0.753 & 0.272 & 0.286 & 0.270 & 0.085 & 0.007 & \textbf{-0.052} & 0.367 & 0.790\\
    3-Player & 0.703 & 0.287 & 0.296 & 0.286 & 0.080 & 0.004 & 0.017 & 0.472 & 0.831 \\
    CAR & - & 0.314 & 0.281 & 0.280 & 0.184 & 0.133 & - & - & - \\
    A2R & 0.654 & 0.268 & 0.287 & 0.264 & 0.084 & 0.008 & 0.128 & 0.338 & 0.581 \\
    A2R-Noise & 0.618 & 0.258 & 0.275 & 0.249 & 0.081 & 0.005 & 0.211 & 0.408 & 0.553\\
    RTP & \textbf{0.777} & \textbf{0.445} & \textbf{0.362} & \textbf{0.416} & \textbf{0.226} & \textbf{0.194} & 0.088 & \textbf{0.697} & \textbf{1.197} \\
    \bottomrule
  \end{tabular}
  \caption{
    Results on the INAS dataset, divided into groups of standard-baselines, post-hoc explainability methods and rationalized neural networks. Best scores per metric are bold, second best are underlined.
  }\label{tab:bio}
  \vspace{-10px}
\end{table*}

\subsection{Advantages}
Our proposed scheme solves all issues discussed in Section \ref{sec:issues}, and is (to our knowledge) the only method to do so. The main advantages lie in the fully differentiable formulation that does not require sampling and gradient approximation, the fact that class-wise rationales are created, and especially in the fact that the classifier is trained on the unaltered inputs instead of on the rationalized variants. This last point ensures that the rationales do not dictate the classification result, but instead explain the actual classification made by the classifier, and therefore completely bypasses issues like interlocking or the dominant selector, thus enforcing high rationale faithfulness. It also means that no degradation in classification performance is to be expected.

\section{Experiments}
We evaluate our method with regards to matching human evidence annotations for text classifications, as well as with regards to the faithfulness of the explanations with regards to the classifier. For details regarding the model and the training and prediction procedures, see Appendix \ref{app:exp_details}.

\subsection{Datasets}
In our evaluation, we use two text classification datasets with span-level evidence annotations, each posing different challenges. The first is the movie review dataset \cite{zaidan2007movies}, containing 2000 reviews with sentiment labels (\textit{positive} or \textit{negative}) and span-level evidence annotations. For this dataset, \citet{DeYoung2020eraser} provided more comprehensive rationales for the test split, which we use in our evaluation. Since class labels are mutually exclusive, this dataset allows models to perform optimally even without class-wise rationales. Additionally, this dataset enables the optimal assessment of the agreement between predicted rationales and the human annotations, since the simplicity of the classification task eliminates the lack of understanding of the inputs as a cause for mismatches.

As for a more challenging classification task, we use the INAS dataset \cite{brinner2022linking}, consisting of 954 scientific paper titles and abstracts from the domain of invasion biology together with labels indicating which hypothesis (from a set of 10 common hypotheses in the field) is addressed in each paper. In a subsequent study, \citet{brinner2024ratio} provided span-level evidence annotations for 750 of the samples. Since some samples belong to multiple classes, optimal performance on this dataset requires class-wise rationalization. Additionally, the more challenging nature of the classification task can highlight degraded classification performance of rationalized models.

\subsection{Evaluation Metrics}
We evaluate the consistency with human annotations on token-level and span-level as done in \cite{brinner2024ratio}, and evaluate the faithfulness of rationales with respect to the classifier as done in \cite{brinner2023MaRC}.

\textbf{Token-Level Evaluation} To evaluate agreement with human rationales at the token level, we use the area under the precision-recall curve (\textit{AUC-PR}). We also assess the token-level F1 score (\textit{Token-F1}), which requires binary predictions. This is done by selecting the highest-scoring $p$ percent of tokens as positive predictions, calculating the standard F1 score, and averaging over 19 values of $p$ (5, 10, ..., 95). For a better absolute assessment of prediction quality, we use the discrete token-level F1 score (\textit{D-Token-F1}), where the top $k$ tokens are selected as the binary rationale and are evaluated with the F1 score, with $k$ being the number of tokens annotated in the corresponding ground truth.

\textbf{Span-Level Evaluation} We also evaluate the quality of predicted spans of text, defined as consecutive words selected as part of the rationale after binary thresholding. The span-level IoU-F1 score (\textit{IoU-F1}) is calculated by determining spans in both the binary rationale prediction and the ground-truth annotation, calculating the IoU for all span pairs, and selecting the maximum IoU value for each predicted and annotated span. This effectively specifies, how well any predicted or ground-truth span overlaps with a span from the other group. IoU-precision and IoU-recall are then defined as the averages of these maximum IoU values for predicted and ground-truth spans, respectively, from which the usual F1 score can be calculated. The holistic IoU-F1 score is then obtained by averaging over the same 19 discrete token selections used for the token-level F1 score. The discrete IoU-F1 score (\textit{D-IoU-F1}) is again calculated by selecting the top-scoring tokens to match the number of tokens specified in the ground-truth annotation.

\textbf{Faithfulness Evaluation} We evaluate rationale faithfulness using scores for sufficiency and comprehensiveness of the predicted rationales. The sufficiency score measures the model's ability to predict the correct label using only the highest-scoring words in the rationale. A lower sufficiency score indicates that fewer tokens are needed for a correct prediction, thus indicating a more faithful rationale:

\vspace{-10px}
\begin{align}
\label{eq:suff}
    \textrm{sufficiency}(x, r) = \frac{1}{19} \sum_{i=1}^{19} M(x) - M(r_i)
\end{align}\vspace{-10px}

The comprehensiveness score is higher if removing the highest-scoring words according to the rationale quickly degrades the model's predictions, again indicating faithful rationales:

\vspace{-10px}
\begin{align}
\label{eq:comp}
    \textrm{comp}(x, r) = \frac{1}{19} \sum_{i=1}^{19} M(x) - M(x \backslash r_i)
\end{align}\vspace{-10px}

In these equations, $x$ denotes the input sample, $r_i$ denotes the $(i\cdot5)\%$ of input tokens with the highest scores according to the rationale, $x \backslash r_i$ denotes the input $x$ with the tokens from $r_i$ removed, and $M(x)$ denotes the probability that model $M$ assigns to the correct class given input $x$. To avoid relying on a single threshold, these scores are calculated by summing over different percentages of rationale tokens used or removed, respectively.

\textbf{Overall Performance}
Ideally, a model should produce rationales that both agree with human rationales and demonstrate faithfulness. We therefore provide an overall performance score (\textit{Perf.}) that sums over the Token-F1, IoU-F1, comprehensiveness and negative sufficiency scores, thus assessing agreement and faithfulness comprehensively.

\begin{table*}[t]
  \centering
  \fontsize{9.5pt}{10.8pt}\selectfont
  \begin{tabular}{lccccccccc}
    \toprule
    Method  & Clf-F1 & AUC-PR & Token-F1 & D-Token-F1 & IoU-F1 & D-IoU-F1& Suff. $\downarrow$ & Comp.$\uparrow$ & Perf. \\
    \midrule
    Random & - & 0.316 & 0.326 & 0.312 & 0.061 & 0.002 & 0.227 & 0.238 & 0.398 \\
    Supervised & 0.980 & 0.670 & 0.514 & 0.626 & 0.144 & 0.169 & 0.001 & 0.638 & 1.295 \\
    \midrule
    MaRC & \underline{0.965} & 0.428 & 0.404 & 0.423 & \underline{0.181} & \underline{0.118} & 0.036 & 0.478 & 1.027 \\
    Occlusion & \underline{0.965} & 0.409 & 0.367 & 0.377 & 0.151 & 0.079 & -0.021 & 0.569 & 1.108 \\
    Int. Grads & \underline{0.965} & 0.376 & 0.358 & 0.371 & 0.067 & 0.009 & 0.049 & 0.484 & 0.860 \\
    LIME & \underline{0.965} & 0.379 & 0.361 & 0.369 & 0.076 & 0.014 & 0.005 & 0.603 & 1.035 \\
    Shapley & \underline{0.965} & 0.442 & 0.390 & 0.426 & 0.082 & 0.020 & \textbf{-0.029} & \underline{0.827} & \underline{1.328} \\
    L2E-MaRC & \underline{0.965} & \underline{0.565} & \underline{0.460} & \underline{0.534} & 0.126 & 0.104 & -0.016 & 0.652 & 1.254 \\
    \midrule
    2-Player & 0.930 & 0.516 & 0.449 & 0.508 & 0.113 & 0.066 & \underline{-0.024} & 0.210 & 0.796 \\
    3-Player & 0.955 & 0.458 & 0.422 & 0.465 & 0.089 & 0.023 & 0.003 & 0.354 & 0.862 \\
    CAR & - & 0.384 & 0.364 & 0.376 & 0.078 & 0.013 & - & - & - \\
    A2R & 0.955 & 0.474 & 0.433 & 0.486 & 0.111 & 0.046 & 0.109 & 0.320 & 0.755 \\
    A2R-Noise & 0.950 & 0.483 & 0.440 & 0.492 & 0.107 & 0.044 & 0.005 & 0.338 & 0.880 \\
    RTP & \textbf{0.975} & \textbf{0.567} & \textbf{0.466} & \textbf{0.544} & \textbf{0.203} & \textbf{0.195} & \textbf{-0.029} & \textbf{0.851} & \textbf{1.549} \\
    \bottomrule
  \end{tabular}
  \caption{
    Results on the movie reviews dataset, divided into groups of standard-baselines, post-hoc explainability methods and rationalized neural networks. Best scores per metric are bold, second best are underlined.
  }\label{tab:movies}
  \vspace{-10px}
\end{table*}

\subsection{Baseline Methods}
We compare our rationalized transformer predictor (RTP) against other rationalized classifiers, which are a two-player game as proposed by \citet{lei2016rational}, a three-player structure with complement predictor \cite{yu2019rethink}, the CAR framework for class-wise rationale generation \cite{chang2019classwise}, and the A2R method \cite{yu2021interlock} as well as an extension to it using noise injection \cite{storek2023noise}. We also compare post-hoc explainability methods that are applied to a standard classifier, which includes MaRC \cite{brinner2023MaRC}, Occlusion \cite{zeiler2014vis}, Integrated Gradients \cite{sundararajan2017intgr}, LIME \cite{riberio2016lime}, Shapley value sampling \cite{castro2009shapley}, as well as a neural network predictor trained on MaRC rationales (L2E-MaRC, \citet{situ2021L2E}). Finally, we report results for a supervised model trained on rationale annotations and a random predictor as additional baselines. For a more detailed overview, see Appendix \ref{app:baseline_methods}.

\section{Results}
\label{sec:results}

The results for the evaluation on the INAS dataset and the movie review dataset are displayed in Table \ref{tab:bio} and Table \ref{tab:movies}, respectively. Exemplary predictions are displayed in Figure \ref{fig:preiction}, with further examples being included in Appendix \ref{app:examples}.

\subsection{Classification Performance}
On both the INAS and movie review datasets, the RTP demonstrates state-of-the-art classification performance, surpassing even the standard classifier. While this pattern is consistent across both datasets, we refrain from assuming a general improvement in classification performance. Instead, we attribute the observed gains to variations in training runs, which are particularly common for the INAS dataset, as reported by \citet{brinner2022linking}. Notably, all other rationalized classifiers consistently exhibit reduced classification performance, thus establishing the RTP as the best-performing model of its kind.

\subsection{Token-Level Performance}
For token-level rationale evaluations on both the INAS and movie review datasets, our RTP method achieves superior performance across AUC-PR, token-F1, and discrete token-F1 metrics. Among competing methods, only L2E-Marc - a neural network trained to predict rationales generated by the MaRC method - consistently approaches the RTP's performance. A detailed discussion of the strong performance of these exact two methods is presented in Section \ref{sec:analysis}.

Notably, the RTP and other post-hoc methods particularly benefit from the ability to predict class-wise rationales on the INAS dataset. Among rationalized neural networks, only the CAR method shares this capability. This distinction is further underscored by the increased performance of class-agnostic rationalized networks on the movie review dataset, where class-wise rationale prediction is irrelevant. On this task, the performance gap with the RTP narrows, and some class-agnostic networks even surpass most post-hoc methods.

The supervised baseline outperforms all methods in token-level predictions, which is to be expected, given that the weakly supervised methods did not receive any supervision regarding the desired rationale output. However, the RTP achieves results that closely approach the supervised baseline across several metrics, demonstrating that in the absence of labeled rationales, the weakly supervised framework offers an effective alternative.

\subsection{Span-Level Performance}
On both datasets, our RTP method is consistently the best performing method with regards to the IoU-F1 and the discrete IoU-F1 scores. Compared to the other rationalized methods this is to be expected, since the RTP has been explicitly designed to extract longer spans of text as rationales. In contrast, other rationalized predictors often rely on a total variation regularizer in their optimization objective, which we found to be ineffective since increasing its strength quickly leads to degenerate solutions with either all or none of the words being selected. This highlights the importance of using the MaRC mask parameterization that reliably leads to the desired results. Notably, the RTP comes close to the supervised method on the INAS dataset without any supervision regarding the usual form of human annotations. On the movie review dataset, the RTP even outperforms the supervised method due to the rationales from the test set being more extensive, thus causing a mismatch between training and test data distributions. This shows that even if slightly inaccurate training data is available for a given task, using a weakly supervised method instead might be preferable.

\subsection{Faithfulness Results}

Our RTP method demonstrates competitive performance with regards to rationale sufficiency, achieving state-of-the-art performance on the movie review dataset, while delivering solid but comparatively weaker results on the INAS dataset. We found that assigning high scores to few important words distributed throughout the whole input is a great strategy for achieving good sufficiency scores (as done, for example, by the Shapley value sampling method), since the model can quickly recognize the correct label from these few highly indicative words. Our RTP model still performs well despite being explicitly discouraged from pursuing this strategy, indicating that our optimization objective is reasonable for generating faithful rationales.

For comprehensiveness, the RTP attains state-of-the-art results on both the INAS and movie review datasets, with only the Shapley value sampling method being close across both tasks and most other methods being outperformed by a large margin.
In general, good faithfulness scores for Shapley value sampling are to be expected, since its objective for scoring input tokens aligns closely with the evaluation measures for faithfulness. Having our method match or surpass the scores of this method shows the exceptional ability of our learning framework for teaching the model to use the rationales to correctly report on its own reasoning.

Overall, our RTP method compares favorably to other rationalized neural networks, since it optimizes its rationales to actually explain the classification, while other methods might, for example, suffer from issues like a dominant predictor that already dictates a specific label. One additional downside of other rationalized models it that they train the predictor on the rationales, which leads to a constant mismatch between the current predictor and the predictor that the rationales have been trained to explain.

Another important insight is, that post-hoc explanation methods do not offer an advantage over the rationales generated by the RTP. Considering, that post-hoc explainers outperform other rationalized networks with respect to faithfulness of the explanations, our method is the first all-in-one method that offers both predictions and rationales with state-of-the-art faithfulness in a single forward pass.

\subsection{Overall Performance}
As discussed, the RTP achieves state-of-the-art results in agreement with human rationales and rationale faithfulness, resulting in dominant scores for overall performance (\textit{Perf.}) on both tasks. In comparison, other rationalized neural networks fall significantly short, with only few post-hoc methods coming somewhat close. These methods have the downside of a substantially higher computational cost in producing a rationale, with, for example, MaRC and Shapley value sampling requiring hundreds of forward passes to create a single rationale.
\section{Discussion}
\label{sec:analysis}

The RTP model demonstrated strong performance across all evaluated metrics. Comparing it specifically to the MaRC method, it outperformed it in every metric related to measuring agreement with human annotations and most faithfulness metrics. This is notable since the RTP can be seen as a neural network parameterized version of the MaRC approach, which originally optimized mask parameters for each sample individually instead of training a neural network to directly predict them from the input. Another well-performing method, especially with regards to token-level evaluation, is the L2E-MaRC method. The L2E framework \cite{situ2021L2E} trains a neural network on pre-calculated rationales created by a post-hoc explainer. Even though it only saw rationales produced by the MaRC method, it manages to outperform it on all metrics measuring token-level agreement with human rationales. These two results indicate, that training to explain many different samples leads to better generalization, which we attribute to reduced overfitting to one specific input. This effect is crucial for the RTP, since it performs input optimization with respect to specific neural network outputs, which has been shown to generally lead to unexpected and uninterpretable artifacts \cite{simonyan2013deep}. The MaRC method successfully mitigated this issue by combining constrained optimization with heavy regularization, but artifacts (i.e., unexpected spans included in the rationale) are still to be expected. In the case of the RTP, training on many samples further reduces this issue, since these unwanted gradient signals will generally not match between different samples, so that the neural network mainly adapts to the wanted signal that is consistent within larger parts of the training set, and that correspond to features that are generally indicative of the respective class.

\section{Conclusion}
We presented a new method for training a rationalized transformer predictor and demonstrated its strong performance on two natural language processing benchmarks. Since our proposed training scheme is not invasive to the general training process and does not produce significant overhead during prediction, we believe that this approach has the potential to facilitate wider adoption and availability of rationalized predictors. Given that transformers are widely used in other modalities like images \cite{dosovitskiy2021image} and audio data \cite{verma2021audio}, we hypothesize that our approach can be extended to these modalities and potentially lead to results of similar quality.

\section{Limitations}

While our method for rationalization generally does not interfere with the training of the prediction module and does not produce notable overhead during prediction, it nevertheless increases the computational cost of model training due to a second forward pass through the model, as well as through more training epochs being required due to slower convergence of rationale training compared to the classification component.

Additionally, the exact form of the produced rationales depends on the models inner working, so that generally a high overlap with human rationales is not guaranteed in cases where the model's reasoning and human reasoning differ.

Finally, while having access to word-level rationale scores is generally helpful, this does not equate to a complete description of the model's inner workings and the actual reasoning process, which most likely is impossible to represent in such a simple form.

\section*{Acknowledgements}
This work was funded by Deutsche Forschungsgemeinschaft DFG (project
number 455913229; T.H., M.B., J.M.J., B.K-R, S.Z.).

\bibliography{main}

\appendix

\section{Experimental Details}
\label{app:exp_details}

The code for our experiments is available at \url{https://github.com/inas-argumentation/RationalizedTransformerPredictor}.

\subsection{Baseline Methods}
\label{app:baseline_methods}

We evaluate our rationalized transformer predictor against a variety of baseline methods that pursuit different strategies for rationalizing predictions. This section provides a general overview, with many model and training details being discussed in Appendix \ref{app:model_details}. The first group are rationalized neural networks, that learn to create rationales from sample-level labels alone. We evaluate the performance of the following methods:
\begin{itemize}
    \itemsep-3px
    \item \textbf{2-Player}: A two-player structure using a rationale extractor and a predictor as proposed by \citet{lei2016rational}. We used an own implementation, since the original work did not use transformers.
    \item \textbf{3-Player}: A three-player structure using a rationale extractor, a predictor and a complement predictor as proposed by \citet{yu2019rethink}. We used an own implementation, since the original work did not use transformers.
    \item \textbf{CAR}: The CAR framework for creating class-wise rationales \cite{chang2019classwise}. We use an own implementation, since the original work did not use transformers. Additionally, we use more extensive parameter sharing, as the original work use a separate rationale predictor for each class, which is impracticable especially for the 10-class classification problem on the INAS dataset. Therefore, a single BERT model predicts rationales for each class at the same time, while a second BERT model acts as the single predictor.
    \item \textbf{A2R}: The A2R framework as proposed by \cite{yu2021interlock}. We use the implementation of \cite{storek2023noise}, who created an implementation relying on BERT models, which, according to their evaluation, outperformed the original implementation that relies on GRUs.
    \item \textbf{A2R-Noise}: The A2R framework with additional noise injection as proposed by \cite{storek2023noise}. We use the implementation provided by the original study.
\end{itemize}

We also evaluated a variety of post-hoc explanation methods:

\begin{itemize}
    \itemsep-3px
    \item \textbf{MaRC}: The MaRC method as proposed by \cite{brinner2023MaRC}. We use the updated weight regularizer proposed by \cite{brinner2024ratio}.
	\item \textbf{Occlusion}: The occlusion method as proposed by \citet{zeiler2014vis}. We chose to mask slightly larger spans of $5$ tokens as this produced smoother masks which resulted in higher IoU F1 scores. We use the implementation by \citet{captum}.
	\item \textbf{Int. Grads}: The integrated gradients method \cite{sundararajan2017intgr}. We use the implementation by \cite{captum}.
    \item \textbf{LIME}: The LIME method \cite{riberio2016lime}. We train a linear classifier on scores from 50 function evaluations. In each evaluation, $5-13\%$ of tokens are selected and the thee tokens starting from the chosen token are removed as input perturbation. We use the implementation by \cite{captum}.
    \item \textbf{Shapley}: Shapley value sampling \cite{castro2009shapley}. We perform $25$ feature permutations per sample, and use the implementation by \cite{captum}.
    \item \textbf{L2E-MaRC}: The L2E framework \cite{situ2021L2E}. We use the rationales created by the MaRC method on the training samples. We discretize the rationales into 5 bins and train a classifier on this dataset. For scoring, we predict the bin-probabilities for each word, multiply them by the bin-means and sum over the resulting values to get a single, continuous score for each word.
\end{itemize}

We also evaluate two further baselines: A random baseline that predicts random scores for each input token, and a supervised method that is trained to perform a binary prediction on each individual token from the input.

\subsection{Model and Training Details}
\label{app:model_details}

\textbf{Base Models} For the movie review experiment, we use bert-base-uncased \cite{devlin2019bert} as base model to stay consistent with previous work and the be able to use existing code bases to ensure implementational accuracy. For the INAS dataset, we use PubMedBERT-base-uncased \cite{gu2021pubmedbert}, since it shows strong performance on the standard classification task for this dataset \cite{brinner2022linking}. These base classifiers are used for all baseline methods, and for all parts of the pipelines (encoder, predictor, base classifier, etc.).

\textbf{Input Processing} During training, samples that exceed the 510 token limit for BERT models were split into multiple segments, and one segment was chosen randomly for this model update. For the evaluation, we again split each sample into smaller parts that adhere to the token limit and that overlap for 100 tokens. Scores were predicted for each split separately and linearly blended afterwards.

\textbf{Model Selection} During training, evaluations on the validation set were performed after each epoch, and the best-performing version of the model was selected for testing. On the INAS dataset, the agreement of the predicted rationales with the human annotations was evaluated after each epoch, and the mean of all five scores (AUC-PR, Token-F1, D-Token-F1, IoU-F1, D-IoU-F1) as well as the classification performance was taken as performance indicator. On the movie dataset, this procedure was not possible, since the data distribution of the validation and test samples is different, meaning that validation results are not a good indicator for performance on the test set. Especially the span-level evaluation scores were unsuitable, since much shorter spans were annotated on the validation set. We chose to use the AUC-PR as performance measure, since it still indicates the models ability to generally recognize useful words, which was again combined with classification f1 as performance indicator.

\subsection{RTP Objective Details}
\label{app:obj_details}
We use two regularizers in the optimization objective for our rationalized transformer predictor. The first is a sparsity regularizer that ensures that only a subset of tokens is selected as rationale:
\begin{align*}
    {\Omega_{\lambda}} = &\sum_{c \in \textbf{y}} \alpha_1 \cdot \textrm{mean}(\textbf{m}^c)^2 + \alpha_2 \cdot \textrm{mean}(\textbf{m}^c) \\
    + &\sum_{c \notin \textbf{y}} \alpha_3 \cdot \textrm{mean}(\textbf{m}^c)^2 + \alpha_4 \cdot \textrm{mean}(\textbf{m}^c)
\end{align*}
In summary, we perform L1 and L2 regularization on the mask means for the masks of the ground truth classes and non-ground-truth classes. In our experiments, we used $\alpha_1=0.2$ and  $\alpha_3=0.05$, meaning that regularization for masks of incorrect classes is weaker. We chose this setting, since for incorrect classes there is no signal that forces words the be unmasked, so that less strong regularization is required. The L1 parameters are set to rather low values of $\alpha_2=\alpha_4=0.001$.

The smoothness regularizer has the following form:
\begin{align*}
    {\Omega_{\sigma}} = \beta_1 \cdot \sum_{c \in \mathcal{Y}} \textrm{mean}((\boldsymbol{\sigma}^c - \beta_2)^2)
\end{align*}
Here, the inner subtraction is meant to be element-wise, so that we regularize each individual sigma value towards a value of $\beta_2$. The actual hyperparameters used are $\beta_1=0.02$ and $\beta_2=3$.

Finally, we use individual weights for each major component of the optimization objective (Equation \ref{eq:objective}):
\begin{align*}
\begin{split}
    \underset{M}{\textrm{arg\,min}}&\:\:\: \gamma_1 \cdot \textbf{CE}(M(\textbf{x}), \textbf{y})\\ &+ \sum_{c\in\textbf{y}}\left[\gamma_2 \cdot \mathcal{L}^c + \gamma_3 \cdot \mathcal{L}^{\overline{c}}\right]\\ &+ \gamma_4 \cdot \Omega_{\lambda} \\ &+ \gamma_5 \cdot \Omega_\sigma
\end{split}
\end{align*}
These values are set to $\gamma_1=2$, $\gamma_2=5$, $\gamma_3=5$, $\gamma_4=3$ and $\gamma_5=3$.

\subsection{Evaluation Post-Processing}
In the INAS dataset, rationales do not cross sentence boundaries. For that reason, we opted to employ a post-processing step that uses SciSpacy \cite{neumann2019scispacy} to split each abstract into sentences, and set the rationale score of the last token in each sentence (that corresponds to punctuation) to 0. This is done for all methods and generally lead to a slight improvement in agreement scores.

\onecolumn
\section{Examples}
\label{app:examples}

\subsection{Movie Reviews Dataset}
\begin{figure}[H]
    \centering
     \includegraphics[width=\textwidth]{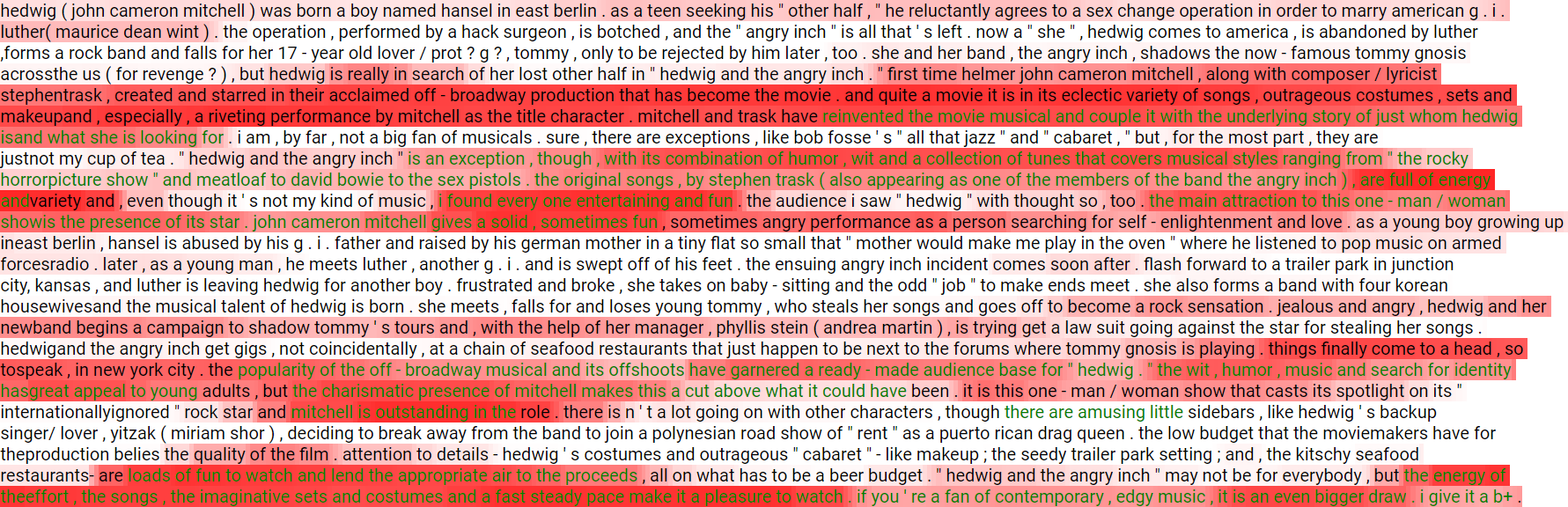}
    \caption{An exemplary output of the RTP for a positive review from the movie reviews dataset. Green text indicates the ground-truth annotations.}
\end{figure}
\begin{figure}[H]
    \centering
     \includegraphics[width=\textwidth]{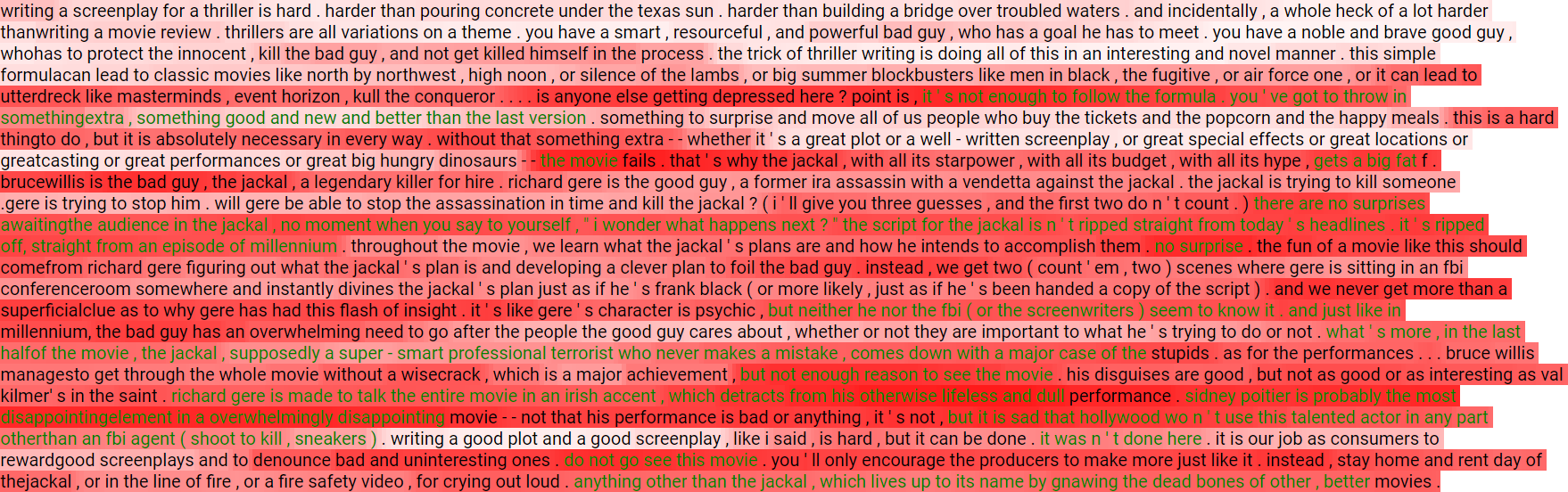}
    \caption{An exemplary output of the RTP for a negative review from the movie reviews dataset. Green text indicates the ground-truth annotations.}
\end{figure}
\subsection{INAS Dataset}
\begin{figure}[H]
    \centering
     \includegraphics[width=\textwidth]{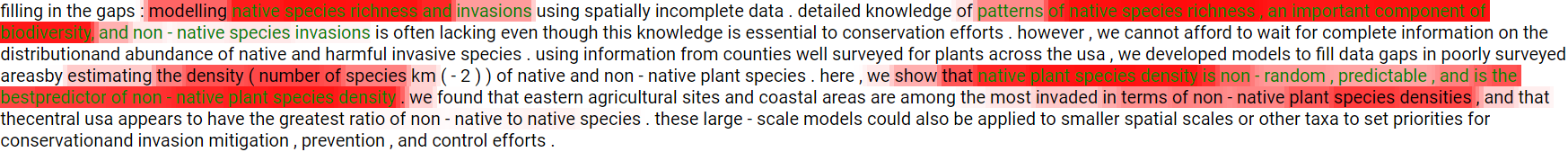}
    \caption{An exemplary output of the RTP for an abstract by \citet{bio1}, which is included in the INAS dataset. The rationale was created for the \textit{Biotic Resistance Hypothesis} label, with green spans indicating the ground-truth annotations.}
\end{figure}
\begin{figure}[H]
    \centering
     \includegraphics[width=\textwidth]{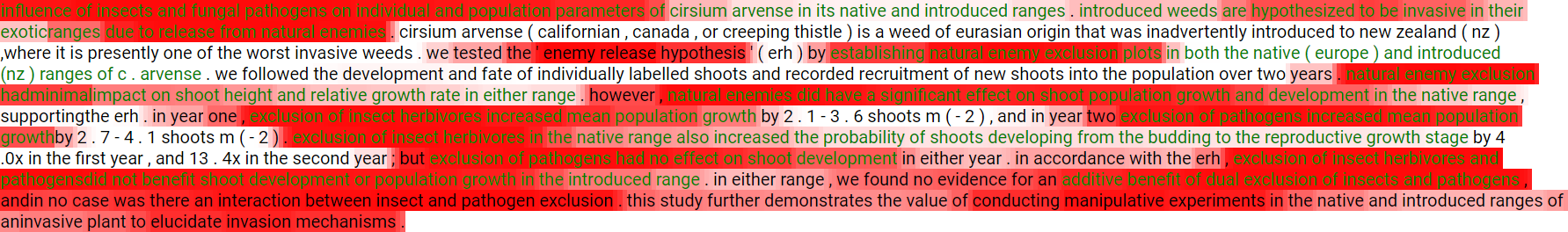}
    \caption{An exemplary output of the RTP for an abstract by \citet{bio2}, which is included in the INAS dataset. The rationale was created for the \textit{Enemy Release Hypothesis} label, with green spans indicating the ground-truth annotations.}
\end{figure}
\end{document}